# Coverage Path Planning for Redundant Manipulators using Generalized Spanning Trees

Raksi Kopo, Kostas J. Kyriakopoulos

***Abstract*— Surface coverage with task-redundant manipulators is challenging because each surface point may admit multiple inverse kinematics (IK) solutions, and configuration choices strongly affect motion quality. This paper extends the classical Spanning Tree Coverage (STC) method to redundant manipulators through offline and online Joint Spanning Tree Coverage (JSTC) algorithms. Offline JSTC samples multiple Inverse Kinematics (IK) solutions per grid cell and formulates the problem as a Generalized Minimum Spanning Tree (GMST), selecting one configuration per cell and tracing the resulting tree to obtain a non-revisiting coverage path. Online JSTC incrementally expands and backtracks a spanning tree with feasibility and cost evaluation while handling dynamic grid updates. Simulation results show that offline JSTC reduces computation time, reconfigurations, and joint motion compared to other methods, while online JSTC achieves fast per-step planning in dynamic scenarios.**

## I. Introduction

Surface coverage tasks, such as cleaning and polishing, require a robot to sweep a tool over a surface while respecting kinematic constraints. For task-redundant manipulators, each surface point may admit multiple Inverse Kinematics (IK) solutions, and the choice of configuration strongly affects motion quality. When the robot's workspace and the surface are partially known or dynamically changing, coverage must be performed online.

In this work, we extend a classical efficient coverage path planning (CPP) algorithm for 2D mobile robots, the Spanning Tree Coverage (STC) [1] method, to surface coverage with task-redundant manipulators. STC operates on a connected grid representation of the surface. Each grid cell is called a mega-cell and is decomposed into its four sub-cells. The STC method outputs a non-revisiting coverage tour that covers every sub-cell of every completely free mega-cell by building a spanning tree of the grid and then tracing the sub-cells around it. The spanning tree can be constructed offline using Depth First Search (DFS), or by solving a Minimum Spanning Tree (MST) problem [2]. While MST is typically more computationally expensive than DFS, it optimizes a desired user-defined edge-cost objective. In the online version, the spanning tree is constructed online using DFS in $O(N)$ time.

We extend STC to task-redundant manipulators in both offline and online settings, resulting in offline Joint Spanning Tree Coverage (JSTC) and online JSTC. In the offline version, the problem is modeled as a Generalized Minimum Spanning Tree Problem (GMSTP) [3]: multiple IK solutions are sampled for each grid cell, and a graph is built by connecting IK solutions of neighboring cells with edges. Those edges are assigned a weight based on the cost of the motion connecting them. A minimum-cost tree is built in the robot's configuration space that covers every grid cell by choosing one IK solution from each. For each sub-cell, we find IK solutions near its mega-cell IK solution using an optimization-based solver. Finally, the sub-cells of the tree are traced as in the original STC to find the configuration space coverage path. In the online version, the tree is expanded using DFS and feasibility checking: at each tree edge extension step, we identify the unvisited neighboring cells of the current cell that the robot can reach from its current configuration and choose the one with the lowest path cost. If no feasible extension is available, the robot backtracks the tree branch to the last cell with a feasible extension available. Due to updates in $G$, a tree may be decomposed into multiple components. In that case, each component is treated as a separate tree to be covered if it still has unvisited subcells. To evaluate the offline version of our algorithm, we compare it with the related work on coverage with redundant manipulators, Joint Generalized Traveling Salesman Problem (JGTSP) [4] and hierarchical JGTSP (HJGTSP) [5], in a benchmark task in simulation. For the online version of our algorithm, we evaluate its effectiveness in several characteristic scenarios in simulation.

The rest of this work is organized as follows: In Section II, we provide a literature review of online CPP methods and methods for manipulator-based surface coverage. In Section III, we describe the offline JSTC algorithm and in Section IV, the online JSTC algorithm. Next, in Section V, we evaluate our algorithms in simulation. Finally, in Section VI we provide concluding remarks and future work directions.

## II. Literature Review

CPP finds many applications in autonomous cleaning [6], manufacturing [7], and structure inspection and maintenance [8]. CPP problems can be categorized according to the type of robot considered (mobile or manipulator), the number of robots involved (single- or multi-robot coverage) [9], the characteristics of the environment (2D or 3D, flat or curved, known or unknown, static or dynamic), and the optimization objective (e.g., minimizing energy consumption, number of turns, overlap, or manipulator reconfigurations) [5]. A comprehensive review of CPP methods can be found in [10].

The authors are with the Center for AI & Robotics (CAIR) and the Electrical Eng. Program, Engineering Division, New York University Abu Dhabi, Abu Dhabi, United Arab Emirates. E-mails: {rk4585,kkyria}@nyu.edu.

### A. Online 2D CPP methods

Online CPP methods plan a coverage path in an environment that is not fully known in advance but is gradually discovered or changing with time due to dynamic obstacles, the introduction of new obstacles, and/or areas to be covered. In such cases, the robot should be able to efficiently compute the coverage path to cover a newly discovered area and avoid a newly discovered obstacle. A class of probabilistically complete, online CPP methods [11], [12] relies on random walks such as Brownian motion or Lévy flight to cover the environment. Compared to deterministic algorithms, these methods perform worse in terms of overlap and coverage time; however, they require no localization and mapping. Decomposition-based methods [13], [14] split the environment into cells that can be covered by simple back-and-forth motions, then graph traversal methods such as DFS or the Traveling Salesman Problem (TSP) are used to determine the visiting order of the cells. In [15], the authors develop an online version of the Morse decomposition-based coverage method by incrementally constructing the Morse decomposition as the robot explores the environment. In [16], the online STC method is extended to handle dynamic obstacles with various relative velocities with respect to the robot. Finally, the method in [17] works for partially known 2D static environments by replanning parts of offline precomputed coverage paths.

### B. CPP methods for surface coverage with manipulators

The aforementioned methods are primarily designed for mobile robots (or purely task-space end-effector (EE) paths). Surface CPP for manipulators poses unique challenges arising from the manipulator's kinematics. Points on the surface can be covered by zero, one, or multiple robot configurations, depending on the robot's tool type, kinematics, external obstacles, and joint limits. Finding a complete coverage path, i.e., a configuration path that connects one IK solution from each point while avoiding reconfigurations, such as large nullspace joint motion or tool lift-offs, is a challenging problem. In the JGTSP method [4], this problem is formulated as a Generalized Traveling Salesman Problem (GTSP). This method starts by uniformly sampling multiple tool poses that cover the surface and multiple IK solutions for each tool pose. Then, an undirected weighted graph is built by connecting IK solutions from different points with edges. The weight of each edge represents the cost of moving from one of its endpoint IK solutions to the other; reconfigurations are assigned high costs. Solving the associated GTSP for that graph means finding a minimum-cost tour that visits every tool pose, choosing only one IK solution from each. Due to the combinatorial complexity of GTSP with respect to the graph nodes, this method is limited to offline settings, relatively small surfaces, and few IK samples per pose. A hierarchically accelerated variant was proposed in [5], improving scalability and reducing computation time, but still requiring offline computation. Another class of methods [18] for non-redundant manipulators begins by projecting the different disconnected components of the surface-constrained configuration space (SCCS) onto the task surface. For surface points that admit IK solutions belonging to multiple disconnected SCCS components, the algorithm selects the component from which the point should be covered to minimize the transitions between disconnected components (i.e., tool lift-offs).

Despite the extensive literature on online CPP for 2D mobile robots, existing methods do not directly extend to task-redundant manipulators, where each surface point may admit multiple IK solutions and the choice of configuration affects motion quality. Likewise, manipulator-based surface coverage methods such as JGTSP [4], [5] primarily target offline settings and require solving combinatorial optimization problems over many sampled IK solutions, which limits their applicability to online scenarios. To the best of our knowledge, no prior method provides online surface coverage for task-redundant manipulators while explicitly accounting for multiple IK solutions per surface point. This gap motivates the proposed offline and online JSTC algorithms, which retain the efficiency and non-revisiting structure of STC over the reachable subset of cells to be covered while selecting feasible low-cost motions in the robot's configuration space.

The main contributions are twofold: an efficient offline JSTC method for surface CPP with task-redundant manipulators, and an online JSTC extension that handles dynamic surface updates.

## III. Offline JSTC

In this section, we formalize the offline coverage problem for known surfaces and present the proposed offline Joint Spanning Tree Coverage (JSTC) algorithm. An illustrative diagram of the algorithm is shown in Fig. 1.

### A. Problem Formulation

Let $\boldsymbol{\xi} \in \mathbb{R}^k$ be the configuration of a $k$-DoF manipulator equipped with a coverage tool. Let the coverage-effective pose of the tool be described by its position $\boldsymbol{p}_t(\boldsymbol{\xi}) \in \mathbb{R}^3$ and axis direction $\boldsymbol{n}_t(\boldsymbol{\xi}) \in \mathbb{R}^3$. We write

$$\boldsymbol{x}_t(\boldsymbol{\xi}) = [\boldsymbol{p}_t(\boldsymbol{\xi})^T, \boldsymbol{n}_t(\boldsymbol{\xi})^T]^T. \quad (1)$$

Let $G = (C, E)$ be a square grid graph representing the part of the surface to be covered, where $C$ is the set of $N$ cells and $E$ is the set of edges induced by 4-neighborhood connectivity. Following STC terminology, the cells of $G$ are called mega-cells. Each mega-cell is decomposed into four equal sub-cells, giving a finer graph $G_s = (C_s, E_s)$, also with 4-neighborhood connectivity. We denote the four sub-cells of a mega-cell $c \in C$ by $C_s(c) \subset C_s$, and the parent mega-cell of a sub-cell $s \in C_s$ by $\mu(s) \in C$.

For any cell $a \in C \cup C_s$, let $\boldsymbol{p}(a) \in \mathbb{R}^3$ be the position of its center on the surface and $\boldsymbol{n}(a) \in \mathbb{R}^3$ the surface normal at that point. The coverage target of $a$ is defined as

$$\boldsymbol{x}(a) = [\boldsymbol{p}(a)^T, \boldsymbol{n}(a)^T]^T. \quad (2)$$

A configuration $\boldsymbol{\xi}$ is an IK solution of $a$ if $\boldsymbol{x}_t(\boldsymbol{\xi}) = \boldsymbol{x}(a)$. This task constrains the tool position and normal direction, leaving rotation about the tool axis free. Therefore, the manipulator is task-redundant when $k > 5$. This definition corresponds to applications such as surface cleaning or polishing with a cylindrical rotating brush. Other applications can be handled by replacing $\boldsymbol{x}(a)$ and $\boldsymbol{x}_t(\boldsymbol{\xi})$ with analogous task-specific kinematic constraints.

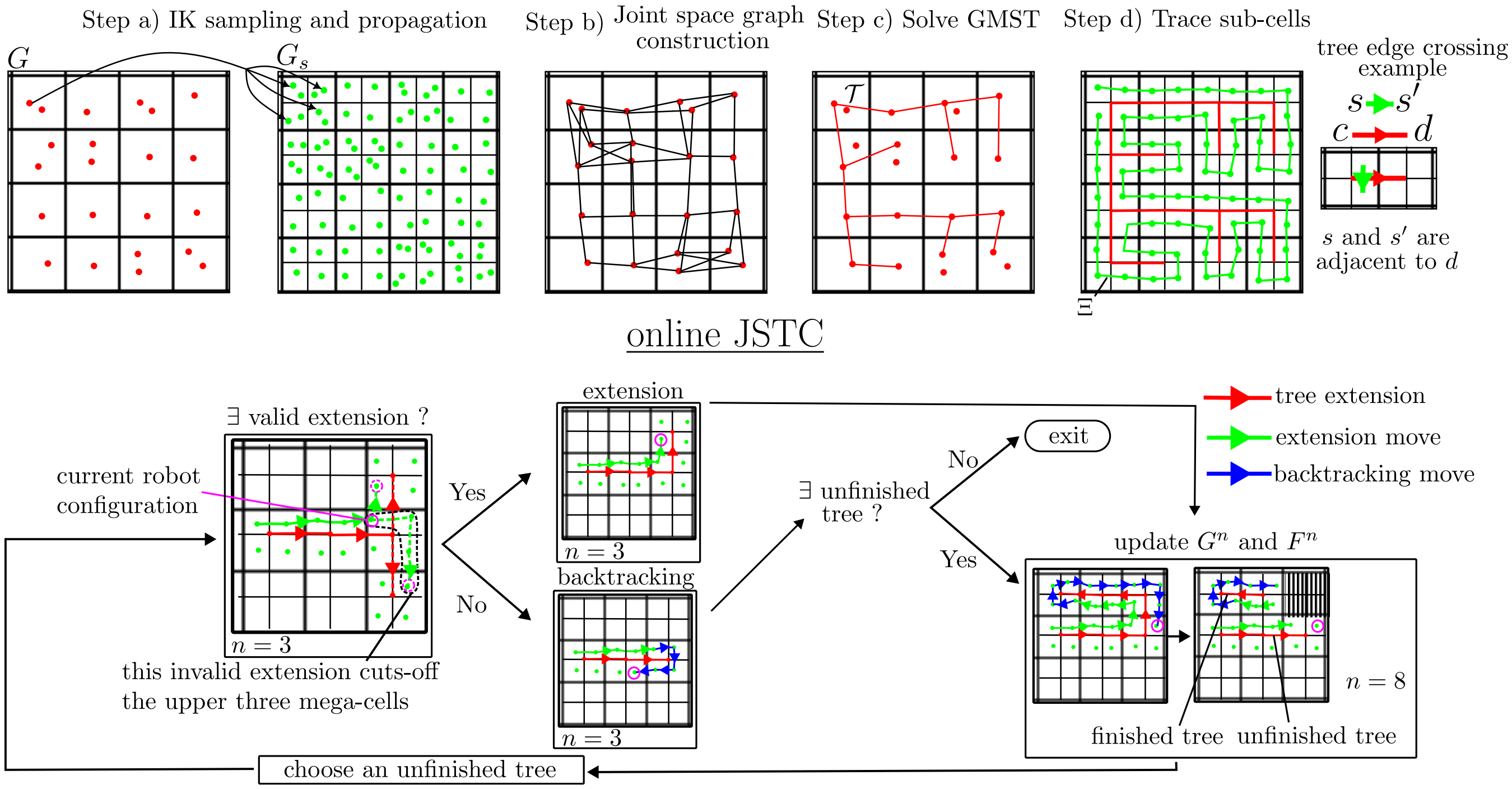

Fig. 1. Illustration of the main steps of the offline and online JSTC algorithms. Mega-cell IK solutions are shown as red dots, whereas sub-cell IK solutions are shown in green. The upper row illustrates the offline pipeline, including IK sampling and propagation, joint-space graph construction, GMST solution, and sub-cell tracing. A tree-edge crossing example is also shown. The lower row illustrates the online procedure, visualising the tree at characteristic steps of the algorithm. During the extension-validity check at step $n = 3$, candidate extensions that would disconnect the remaining unvisited sub-cells are rejected. During the update step at step $n = 8$, the removal of a mega-cell, shown by hatching, splits the spanning tree into two components. One of them has no unvisited sub-cells, thus it is labeled as "finished".

The coverage task is to compute a sequence of sub-cell IK solutions

$$\Xi = \{\boldsymbol{\xi}_{s_1}, \boldsymbol{\xi}_{s_2}, \dots, \boldsymbol{\xi}_{s_{4N}}\}, \tag{3}$$

where each sub-cell $s \in C_s$ is visited once and $\boldsymbol{\xi}_s$ is an IK solution of $s$. The desired path should have low joint motion and few reconfigurations. Instead of directly optimizing the full sub-cell path, offline JSTC first builds a low-cost tree over mega-cell IK solutions and then traces the corresponding sub-cells around this tree, as in STC.

We define a reconfiguration similarly to [5]. Given two IK solutions $\boldsymbol{\xi}_1$ and $\boldsymbol{\xi}_2$ of cells $a_1$ and $a_2$, respectively, the transition between them is called a reconfiguration if any of the following conditions holds:

- $a_1$ and $a_2$ are not neighboring cells;
- $|\boldsymbol{\xi}_1 - \boldsymbol{\xi}_2|_\infty > \tau_1$, where $\tau_1 > 0$ is the maximum allowed joint displacement;
- along the straight-line interpolation in joint space between $\boldsymbol{\xi}_1$ and $\boldsymbol{\xi}_2$, the tool deviates from both endpoint targets by more than $\tau_2 > 0$.

The deviation between two coverage targets $\boldsymbol{x}_1 = [\boldsymbol{p}_1^T, \boldsymbol{n}_1^T]^T$ and $\boldsymbol{x}_2 = [\boldsymbol{p}_2^T, \boldsymbol{n}_2^T]^T$ is

$$\Delta(\boldsymbol{x}_1, \boldsymbol{x}_2) = |\boldsymbol{p}_1 - \boldsymbol{p}_2|_2 + \alpha \arccos(\boldsymbol{n}_1^T \boldsymbol{n}_2), \tag{4}$$

where $\alpha > 0$ is a weighting constant. We denote by $A(\boldsymbol{\xi}_1, \boldsymbol{\xi}_2) \in \{0, 1\}$ the indicator that equals one when the transition is a reconfiguration.

### B. Offline JSTC Algorithm

Offline JSTC consists of four steps: IK sampling and propagation, joint graph construction, GMST solving, and sub-cell tracing. The pseudocode of offline JSTC is given in Algorithm 1.

*a) IK sampling and propagation:* For each mega-cell $c \in C$, we sample $m$ IK solutions

$$\Xi_c^G = \{\boldsymbol{\xi}_1^c, \boldsymbol{\xi}_2^c, \dots, \boldsymbol{\xi}_m^c\}. \tag{5}$$

This is done by uniformly sampling $m$ initial seed configurations $\boldsymbol{\xi}_i^{\text{seed}}$, $i = 1, \dots, m$, within the joint limits and running an optimization-based IK solver for the target $\boldsymbol{x}(c)$. The solver is initialized at $\boldsymbol{\xi}_i^{\text{seed}}$ and minimizes the joint distance from this seed. We write

$$\boldsymbol{\xi}_i^c = \text{IK}(\boldsymbol{x}(c), \boldsymbol{\xi}_i^{\text{seed}}), \quad i = 1, \dots, m. \tag{6}$$

In our implementation, we set $m = 100$ and solve the IK problem using CasADi's IPOPT solver [19]. Similar IK solutions are merged using DBSCAN [20].

Each mega-cell IK solution is then *propagated* to the four sub-cells of the same mega-cell. For every $s \in C_s(c)$, we compute

$$\boldsymbol{\xi}_{i,s}^{c} = \mathrm{IK}(\boldsymbol{x}(s), \boldsymbol{\xi}_i^c), \tag{7}$$

using the mega-cell IK solution as the initial guess. We retain $\boldsymbol{\xi}_i^c$ only if it can be propagated to all four sub-cells of $c$. If no IK solution of a mega-cell can be propagated to all of its sub-cells, that mega-cell is removed from $G$.

*b) Joint graph construction:* We construct a generalized graph whose clusters correspond to mega-cells. Each cluster contains the feasible IK solutions of one mega-cell. Edges are added only between IK solutions of neighboring mega-cells. For two IK solutions $\boldsymbol{\xi}_i^{c_a}$ and $\boldsymbol{\xi}_j^{c_b}$, the edge weight is

$$w\big(\boldsymbol{\xi}_i^{c_a}, \boldsymbol{\xi}_j^{c_b}\big) = \begin{cases} |\boldsymbol{\xi}_i^{c_a} - \boldsymbol{\xi}_j^{c_b}|_2, & \text{if } (c_a, c_b) \in E \ \wedge \ A = 0, \\ B, & \text{if } (c_a, c_b) \in E \ \wedge \ A = 1, \\ \infty, & \text{otherwise}, \end{cases} \tag{8}$$

where $B$ is a large positive constant penalizing reconfigurations between neighbor megacells.

*c) Solving the GMST problem:* We solve a Generalized Minimum Spanning Tree Problem (GMSTP) on the constructed graph. In our implementation, we use the efficient heuristic algorithm of [21], which is based on local search and has complexity $O((mN)^2)$. The output is a generalized spanning tree

$$\mathcal{T} = (C, E_T, \Xi_T, \Xi_{s,T}), \tag{9}$$

where $E_T \subseteq E$ is the selected tree edge set, $\Xi_T$ contains one selected IK solution per mega-cell, and $\Xi_{s,T}$ contains the sub-cell IK solutions propagated from the selected mega-cell IK solutions.

*d) Sub-cell tracing:* The sub-cell path is obtained by tracing around the selected tree, as in STC. Starting from an arbitrary sub-cell, the algorithm repeatedly moves to an unvisited neighboring sub-cell using admissible moves. A move from sub-cell $s$ to sub-cell $s'$ is admissible if $s'$ is a 4-neighbor of $s$ and either: i) $\mu(s) \neq \mu(s')$ and $(\mu(s), \mu(s')) \in E_T$, or ii) $\mu(s) = \mu(s')$ and the move does not cross any tree edge incident to $\mu(s)$. For two neighboring sub-cells inside the same mega-cell $c$, the move crosses an incident tree edge $(c, d) \in E_T$ only when both sub-cells lie on the side of $c$ adjacent to $d$ (see Fig. 1). This rule makes the trace follow the boundary of the tree without crossing tree edges as shown in Fig. 1.

After the ordered sub-cell path

$$\mathcal{P}_s = \{s_1, s_2, \dots, s_{4N}\} \tag{10}$$

is obtained, the final configuration-space coverage path is

$$\Xi = \{\boldsymbol{\xi}_{s_1}, \boldsymbol{\xi}_{s_2}, \dots, \boldsymbol{\xi}_{s_{4N}}\}, \tag{11}$$

where each $\boldsymbol{\xi}_{s_\ell}$ is the propagated IK solution associated with the selected mega-cell IK solution of $\mu(s_\ell)$.

---

**Algorithm 1:** Offline Joint Spanning Tree Coverage

**Input:** Mega-cell graph $G = (C, E)$, sub-cell graph $G_s = (C_s, E_s)$, IK solver
**Output:** Configuration-space coverage path $\Xi$

1 **foreach** $c \in C$ **do**
2   $\Xi_c^G \leftarrow$ sample and merge IK solutions for $\boldsymbol{x}(c)$.; Propagate each $\boldsymbol{\xi}_i^c \in \Xi_c^G$ to all $s \in C_s(c)$.; Remove IK solutions that fail propagation.; **if** $\Xi_c^G = \emptyset$ **then**
3     Remove $c$ from $G$ and remove $C_s(c)$ from $G_s$.;
4   **end**
5 **end**
6 Build the generalized graph with clusters $\{\Xi_c^G\}_{c \in C}$ and edge weights from (8).;
7 Solve the GMSTP to obtain $\mathcal{T} = (C, E_T, \Xi_T, \Xi_{s,T})$.;
8 $\mathcal{P}_s \leftarrow$ trace sub-cells around $\mathcal{T}$ using admissible non-crossing moves.;
9 $\Xi \leftarrow$ propagated IK solutions in $\Xi_{s,T}$ ordered by $\mathcal{P}_s$.;
10 **return** $\Xi$;

---

## IV. Online JSTC

In this section, we present the online version of JSTC for the case where the surface graph is partially known or changes during execution. An illustrative diagram and representative examples are shown in Fig. 1.

### A. Problem Formulation

Let $G^n = (C^n, E^n)$ and $G_s^n = (C_s^n, E_s^n)$ denote the mega-cell and sub-cell graphs at planning step $n$, respectively. A graph update process modifies $G^n$ during execution by adding or removing mega-cells. Added mega-cells correspond to newly discovered regions to be covered, while removed mega-cells correspond to regions that are no longer available or no longer required for coverage.

The goal is to incrementally construct a configuration-space path $\Xi$ that covers every sub-cell of every reachable mega-cell that remains in the updated graph. Unlike offline JSTC, the full graph is not assumed to be fixed in advance, and therefore the generalized spanning tree and the associated sub-cell path must be built and updated online.

The algorithm maintains a forest

$$F^n = \{\mathcal{T}_1^n, \mathcal{T}_2^n, \dots\}, \tag{12}$$

where each tree belongs to a connected component of $G^n$. Each tree is written as

$$\mathcal{T} = (C_T, E_T, \Xi_T, \Xi_{s,T}, \mathcal{P}_{s,T}), \tag{13}$$

where $C_T$ is the set of mega-cells in the tree, $E_T$ is the tree edge set, $\Xi_T$ stores one IK solution per mega-cell, $\Xi_{s,T}$ stores the propagated sub-cell IK solutions, and $\mathcal{P}_{s,T}$ is the ordered sequence of visited sub-cells. We write $\boldsymbol{\xi}_T(c)$ for the IK solution of mega-cell $c$ and $\boldsymbol{\xi}_{s,T}(s)$ for the IK solution of sub-cell $s$. Each tree is rooted at the mega-cell from which it was initialized, and the parent relation induced by $E_T$ is used during backtracking.

A tree is unfinished if it contains unvisited sub-cells. If the robot is already in an unfinished tree, that tree remains active. Otherwise, the nearest unfinished tree is selected, and RRT-Connect [22] is used to connect the current robot configuration to that tree.

### *B. Online JSTC Algorithm*

The online procedure is summarized in Algorithm 2. It builds the tree through two operations: *extension* and *backtracking*. Extension adds a neighboring mega-cell to the active tree, while backtracking traces the tree backward through unvisited sub-cells until a sub-cell of the parent mega-cell is reached.

For compactness, the pseudocode uses the following helper operations. $\mathrm{IKProp}(c, \boldsymbol{\xi}_0)$ computes an IK solution for mega-cell $c$ initialized at $\boldsymbol{\xi}_0$ and propagates it to all sub-cells of $c$ using our IK solver; it returns failure if either the mega-cell IK or any sub-cell propagation fails. $neigh_F(c)$ denotes neighboring mega-cells of $c$ that do not belong to any tree in the current forest. $\mathrm{TraceToNeigh}(s, C'_s, E')$ traces admissible unvisited sub-cells in $\mu(s)$ starting from $s$ until the first unvisited sub-cell in $C'_s$ is reached, without crossing edges in $E'$. $\mathrm{ConnOK}(\mathcal{P}, \mathcal{T}, G_s)$ checks whether the remaining unvisited sub-cells connected to the active tree remain connected after adding path $\mathcal{P}$. Finally, $\mathrm{Append}_\Xi(\mathcal{P}, \Xi_{s,T})$ appends to $\Xi$ the IK solutions stored in $\Xi_{s,T}$ and associated with the sub-cell path $\mathcal{P}$; if a transition is a reconfiguration, the corresponding motion is planned using RRT-Connect.

At initialization, the current mega-cell $c_{\mathrm{cur}}$ and sub-cell $s_{\mathrm{cur}}$ are determined from the robot state. An IK solution is computed for $c_{\mathrm{cur}}$ and propagated to its four sub-cells. A tree is then initialized with this mega-cell, its IK solution, its propagated sub-cell IK solutions, and the initial visited sub-cell.

After each iteration, graph updates are incorporated into the forest. Removed mega-cells are deleted from their trees. If this disconnects a tree, each connected component with unvisited sub-cells is kept as a separate unfinished tree (see Fig. 1). Components whose sub-cells have all been visited are marked as finished. New trees are initialized for uncovered connected components that do not contain an unfinished tree and admit a feasible IK solution.

*a) Extension:* The extension operation is described in Algorithm 3. Given the current mega-cell $c_{\mathrm{cur}}$, a candidate extension is a neighboring mega-cell $v \in neigh_F(c_{\mathrm{cur}})$. The algorithm computes an IK solution for $v$ using $\boldsymbol{\xi}_T(c_{\mathrm{cur}})$ as the initial guess and propagates it to all sub-cells of $v$. If IK propagation succeeds, a tree edge $e = (c_{\mathrm{cur}}, v)$ is formed.

The sub-cell extension path is obtained by tracing admissible unvisited sub-cells until an unvisited sub-cell of $v$ is first encountered. The move is valid only if it does not cross the existing tree edges or the new edge, and if adding the path does not disconnect the remaining unvisited sub-cells, as shown in Fig. 1. Among all valid extensions, the algorithm selects the one with the lowest transition cost.

**Algorithm 2:** Online Joint Spanning Tree Coverage

**Input:** $G^0$, $G^0_s$, IK solver, graph update process
**Output:** Configuration-space coverage path $\Xi$

1 Initialize $c_{\mathrm{cur}}$ and $s_{\mathrm{cur}}$.; $(\boldsymbol{\xi}_T(c_{\mathrm{cur}}), \Xi_{s,T}) \leftarrow \mathrm{IKProp}(c_{\mathrm{cur}}, \boldsymbol{\xi}^{\mathrm{seed}})$.; Initialize $\mathcal{T}$ with $c_{\mathrm{cur}}$, $s_{\mathrm{cur}}$, $\boldsymbol{\xi}_T(c_{\mathrm{cur}})$, and $\Xi_{s,T}$.; Set $F \leftarrow \{\mathcal{T}\}$ and $\Xi \leftarrow \{\boldsymbol{\xi}_{s,T}(s_{\mathrm{cur}})\}$.;
2 **for** $n = 1$ **to** $N_{\max}$ **do**
3 $(G^n, G^n_s, F^n) \leftarrow \mathrm{UpdateF}(G^{n-1}, G^{n-1}_s, F^{n-1})$.; Mark finished trees and initialize new feasible trees in uncovered components.;
4 **if** *F has no unfinished tree* **then**
5 **break**;
6 **end**
7 $\mathcal{T} \leftarrow \mathrm{SelectTree}(F, s_{\mathrm{cur}})$.;
8 **if** *the active tree changes* **then**
9 $s_{\mathrm{cur}} \leftarrow \mathrm{last}(\mathcal{P}_{s,T})$.; $\Xi \leftarrow \Xi \circ \mathrm{RRTConnect}(\mathrm{last}(\Xi), \boldsymbol{\xi}_{s,T}(s_{\mathrm{cur}}))$.; $c_{\mathrm{cur}} \leftarrow \mu(s_{\mathrm{cur}})$.;
10 **end**
11 $\mathcal{V} \leftarrow neigh_F(c_{\mathrm{cur}})$.; $\mathcal{D} \leftarrow \{\mathrm{Extend}(\mathcal{T}, F, G_s, v) : v \in \mathcal{V}\} \setminus \{\emptyset\}$.;
12 **if** $\mathcal{D} \neq \emptyset$ **then**
13 $\Delta\mathcal{T} \leftarrow \arg\min_{\Delta \in \mathcal{D}} J(\Delta)$.; Apply $\Delta\mathcal{T}$ to $\mathcal{T}$.; $\mathcal{P}_{s,T} \leftarrow \mathcal{P}_{s,T} \circ \mathcal{P}_{\mathrm{ext}}$.; $\Xi \leftarrow \Xi \circ \mathrm{Append}_\Xi(\mathcal{P}_{\mathrm{ext}}, \Xi_{s,T})$.;
14 **else**
15 $\mathcal{P}_{\mathrm{back}} \leftarrow \mathrm{Backtrack}(\mathcal{T}, G_s)$.; $\mathcal{P}_{s,T} \leftarrow \mathcal{P}_{s,T} \circ \mathcal{P}_{\mathrm{back}}$.; $\Xi \leftarrow \Xi \circ \mathrm{Append}_\Xi(\mathcal{P}_{\mathrm{back}}, \Xi_{s,T})$.;
16 **end**
17 $s_{\mathrm{cur}} \leftarrow \mathrm{last}(\mathcal{P}_{s,T})$.; $c_{\mathrm{cur}} \leftarrow \mu(s_{\mathrm{cur}})$.;
18 **end**
19 **return** $\Xi$;

*b) Backtracking:* If no valid extension exists, the algorithm backtracks as described in Algorithm 4. Backtracking does not add a new mega-cell. Instead, it traces admissible unvisited sub-cells of the current mega-cell until it reaches the first unvisited sub-cell of the parent mega-cell. If the current mega-cell is the root, the remaining sub-cells of the root are visited and the active tree is marked as finished.

The tracing routines use the same fixed neighbor ordering as offline JSTC. Therefore, "first encountered" refers to the first admissible sub-cell found under this ordering. When the active tree is finished, another unfinished tree is selected from the forest. If the new tree is not directly reachable through the current coverage graph, RRT-Connect is used to connect the robot to it. The algorithm stops when all trees are finished.

## V. SIMULATIONS

In this section, we validate both offline and online JSTC. We compare offline JSTC with the JGTSP [4] and HJGTSP

**Algorithm 3:** Tree Extension

**Input:** $\mathcal{T}$, $F$, $G_s$, candidate mega-cell $v$, IK solver
**Output:** Extension plan $\Delta\mathcal{T}$ or failure

1 $s_{\text{cur}} \leftarrow \text{last}(\mathcal{P}_{s,T})$, $c_{\text{cur}} \leftarrow \mu(s_{\text{cur}})$.;
2 $(\boldsymbol{\xi}_v, \{\boldsymbol{\xi}_s\}_{s\in C_s(v)}) \leftarrow \text{IKProp}(v, \boldsymbol{\xi}_T(c_{\text{cur}}))$.;
3 **if** *IK propagation fails* **then**
4 | **return** failure;
5 **end**
6 $e \leftarrow (c_{\text{cur}}, v)$.;
$\mathcal{P}_{\text{ext}} \leftarrow \text{TraceToNeigh}(s_{\text{cur}}, C_s(v), E_T \cup \{e\})$.;
7 **if** $\mathcal{P}_{\text{ext}} = \emptyset$ **then**
8 | **return** failure;
9 **end**
10 **if** $\neg\text{ConnOK}(\mathcal{P}_{s,T} \cup \mathcal{P}_{\text{ext}}, \mathcal{T}, G_s)$ **then**
11 | **return** failure;
12 **end**
13 $J_{\text{ext}} \leftarrow w(\boldsymbol{\xi}_T(c_{\text{cur}}), \boldsymbol{\xi}_v)$.;
14 **return** $\Delta\mathcal{T} = (v, e, \boldsymbol{\xi}_v, \{\boldsymbol{\xi}_s\}_{s\in C_s(v)}, \mathcal{P}_{\text{ext}}, J_{\text{ext}})$;

**Algorithm 4:** Tree Backtracking

**Input:** Active tree $\mathcal{T}$, sub-cell graph $G_s$
**Output:** Backtracking path segment $\mathcal{P}_{\text{back}}$

1 $s_{\text{cur}} \leftarrow \text{last}(\mathcal{P}_{s,T})$, $c_{\text{cur}} \leftarrow \mu(s_{\text{cur}})$.;
2 **if** $c_{\text{cur}} = \text{root}(\mathcal{T})$ **then**
3 | $\mathcal{P}_{\text{back}} \leftarrow \text{TraceRemaining}(c_{\text{cur}}, E_T)$.; Mark $\mathcal{T}$ as finished.; **return** $\mathcal{P}_{\text{back}}$;
4 **end**
5 $c_p \leftarrow \text{parent}_{\mathcal{T}}(c_{\text{cur}})$.;
$\mathcal{P}_{\text{back}} \leftarrow \text{TraceToNeigh}(s_{\text{cur}}, C_s(c_p), E_T)$.;
6 **return** $\mathcal{P}_{\text{back}}$;

algorithms [5] in the "Scan Floor" benchmark task [5], where a 7 DoF Franka Emika Panda robot uses a 6 DoF rectangular handheld detector to scan the floor around it for potential chemical or radioactive leaks (see Fig. 2 (upper left)). For both JGTSP and HJGTSP, the GLKH heuristic solver [23] is used to solve the associated GTSPs. This solver has empirical computational complexity $O(N_{\text{nodes}}^{2.2})$ where $N_{\text{nodes}}$ is the total number of graph nodes, in this case IK solutions. We test the online JSTC in a table cleaning application where a 7 DoF KUKA LBR iiwa robot equipped with a spherical brush is tasked to clean a table represented by a $8 \times 8$ grid with resolution 15 mega-cells per meter. The tool tip is positioned at the center of the sub-cell under consideration. The rotation about the tool axis is unrestricted, while deviations from the surface normal are bounded within a cone of half-angle, as depicted in Fig. 3a) left. We test our algorithm in five cases: 1) free grid (see Fig. 3a)), 2) an external obstacle (see Fig. 3b)), 3) internal obstacles (i.e. removed grid mega-cells) (see Fig. 4a)), 4) disconnected graph (see Fig. 4b)), 5) removal of tree mega-cells. All tests for all methods were performed on a laptop with an Intel Core i9-11900H processor running at 2.50 GHz, and with 32 GB of RAM. Simulation videos showing the robot's motion are provided[1].

### A. *Offline JSTC test results*

We repeated the test 10 times to account for the randomness of the IK sampling. We used the following metrics to compare the three approaches: 1) mean computation time, 2) mean number of reconfigurations, and 3) mean total joint motion. We also give the number of EE targets, which for our method corresponds to the number of sub-cells of $G$. For JGTSP and HJGTSP, this corresponds to the number of sampled surface points. The results are summarized in Table I along with path errors, while the resulting tool paths are shown in Fig. 2. We notice that our method outperforms JGTSP in terms of computation time, number of reconfigurations, and total joint motion. This can be understood since the JGTSP solver solves a high-dimensional GTSP problem on a graph with at most $N_{\text{targets}} = 674 \cdot m$ nodes, while our algorithm solves the GMST on $G$, which has only $N = 179$ mega-cells, thus $179 \cdot m$ nodes. The significantly lower performance in the number of reconfigurations and joint movement is attributed to the performance of the heuristic GTSP solver, which deteriorates for larger numbers of nodes.

HJGTSP reduces computation time by solving a small higher-level GTSP over $N_{\text{exemplars}} = 24$ representatives and then propagating the selected IK solutions to the remaining $N_{\text{targets}} = 674$ targets. However, the small number of representatives creates larger clusters, so a representative IK solution may not propagate to a good solution for all targets in its cluster. This reduced representativeness can degrade the lower-level GTSP solution, leading to higher total joint motion than our method, which uses a denser higher-level discretization (179 mega-cells vs 24 exemplars). Fig. 2 shows the benchmark scenario, the generated spanning tree, and the tool paths. Due to the grid structure of JSTC, our path consists of horizontal and vertical segments, whereas the mesh-based GTSP methods may include diagonal edges and sharper turns.

### B. *Online JSTC test results*

In case 1) the robot is tasked to cover the grid without obstacles. The resulting path and tree in task space are shown in Fig. 3 a). The black mega-cells correspond to mega-cells the robot cannot reach due to its workspace limits. In case 2), we introduce a spherical obstacle on the surface shown in Fig. 3 b), along with the generated path. The obstacle causes the grid mega-cells shown in red to be unreachable by the tree (i.e. there is no valid extension for them) because of collision. In cases 3 and 4, we introduce internal obstacles by removing the tree mega-cells in black shown in Fig. 4a) and b) along with the generated paths, respectively. For each disconnected region of the grid, our algorithm developed a separate tree as shown in Fig. 4. Finally, in case 5 the robot has performed the green path shown in Fig. 5a) and encounters a grid update that removes the sixth row. This leads to the split of the tree and the path to multiple disconnected components

[1] https://youtu.be/Wh7PqAlNU94

TABLE I

OFFLINE COVERAGE PERFORMANCE COMPARISON ON THE *Scan Floor* BENCHMARK.

| Method | Computation time (s) | # Reconfigurations | Total joint motion (rad) | # EE targets | Pos. error (m) | Rot. error (rad) |
|---|---|---|---|---|---|---|
| JGTSP [4] | $1105.15 \pm 199.01$ | $30.64 \pm 10.91$ | $190.01 \pm 8.70$ | 674 | 9.9e−4 | 7.9e−3 |
| HJGTSP [5] | $206.96 \pm 68.39$ | $1.93 \pm 0.45$ | $129.38 \pm 1.62$ | 674 | 10.0e−4 | 10.0e−3 |
| Offline JSTC (ours) | $264.07 \pm 5.61$ | $4.00 \pm 0.00$ | $83.14 \pm 2.81$ | 716 | 10.0e−4 | 10.0e−3 |

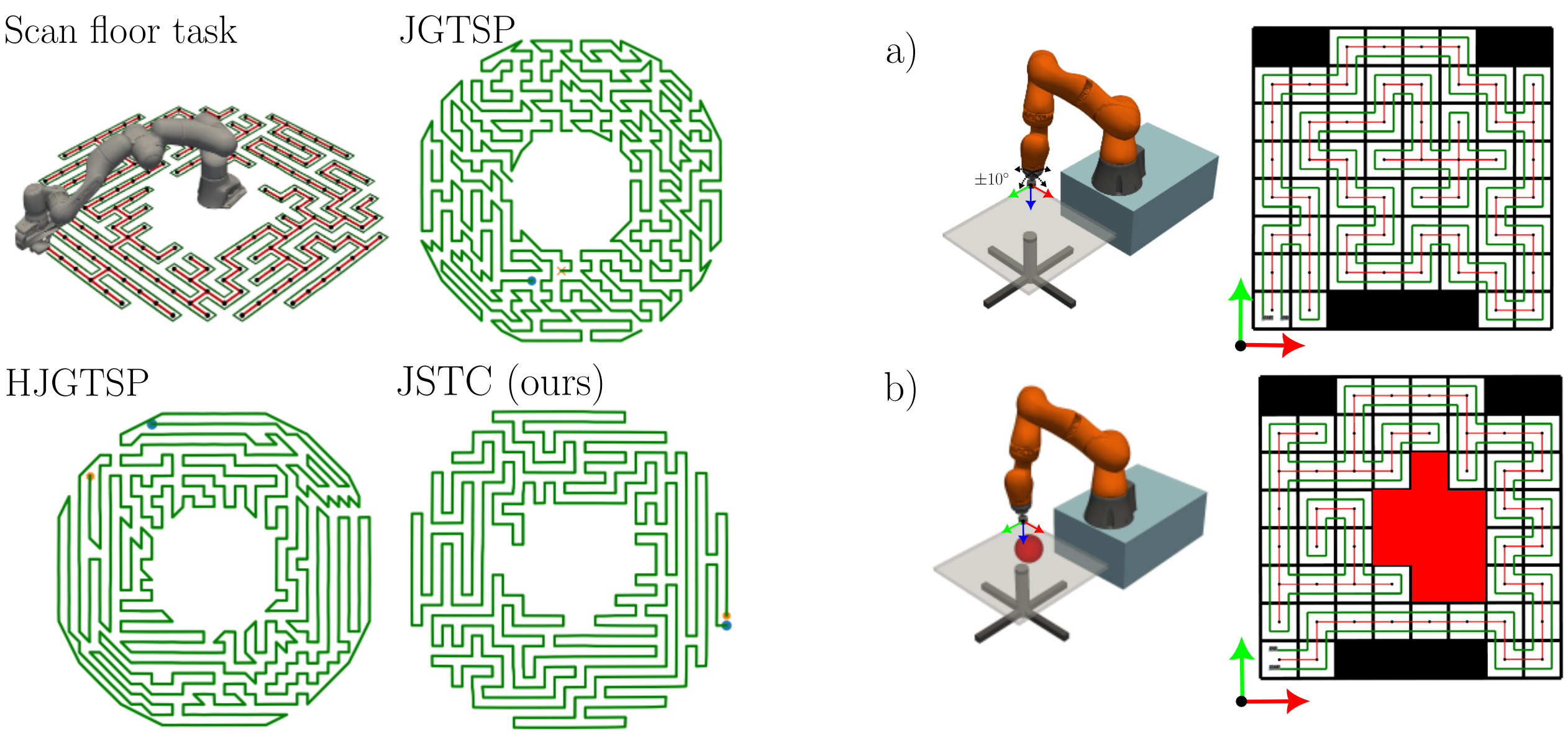

Fig. 2. The figure shows the Scan floor scenario with forward kinematics image of the generated generalized spanning tree (in red) and with black dots the mega-cell centers. The tool paths generated by the offline algorithms are shown in green. The orange blue dot denotes the start, and the orange "x" marker denotes the end of each path.

Fig. 3. Coverage path results for a) free grid and b) external obstacle scenarios (red sphere). The tool coverage path is denoted in green, while the forward kinematic images of the generalized spanning tree are shown in red. The red mega-cells correspond to unreachable mega-cells due to collision with the obstacle, while the black mega-cells correspond to unreachable mega-cells due to workspace limits.

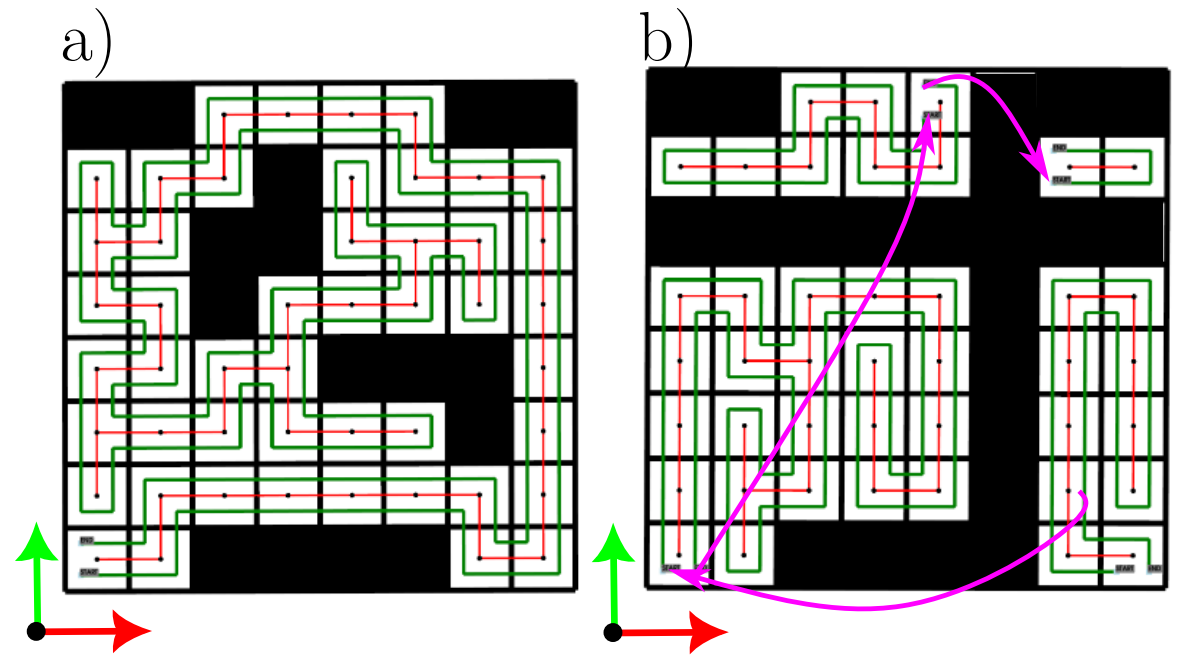

Fig. 4. Coverage path results for internal obstacles in two scenarios a) connected grid b) disconnected grid. The purple arrow denotes transitions between trees that span each disconnected grid component.

shown in Fig. 5 b). The robot completes the path for each such component with the remaining unvisited mega-cells. For each run, we computed the following metrics to assess the performance of our online algorithm 1) average computation time per extension step, 2) average computation time per backtracking step, 3) average computation time per tree and path processing after an update occurs i.e., Steps 3(e), 3(f), 3(g) in Algorithm 2, 4) average joint motion per consecutive IK solutions in the path. Table II shows the results. The average extension step and backtracking step time is low for all cases (less than 70 ms). In case five, where a grid update occurs, the computation time for performing the update of the grid and the associated tree and path splits is also low. The average joint displacement per step shown in the fifth column is acceptable given that our algorithms optimize joint motion only locally compared to the offline JSTC or JGTSP method, which optimized joint motion globally. Overall, the results show that our algorithm can be applied effectively online.

## VI. CONCLUSION

We presented offline and online Joint Spanning Tree Coverage (JSTC), extending Spanning Tree Coverage to surface coverage with task-redundant manipulators. Offline JSTC selects one IK solution per mega-cell through a GMST formulation and traces the resulting tree to obtain a non-revisiting sub-cell coverage path. Online JSTC incrementally extends and backtracks the tree while handling dynamic graph updates. Simulations show that offline JSTC improves computational time motion quality over GTSP-based baselines, while online JSTC achieves low per-step computation time across obstacle and update scenarios. Future work includes experimental validation and extension to curved surfaces.

TABLE II

ONLINE JSTC PERFORMANCE METRICS PER STEP ACROSS FIVE SCENARIOS.

| Case | Extension time (ms) | Backtracking time (ms) | Update proc. time (ms) | Avg. joint motion (rad) |
|---|---|---|---|---|
| 1) Free grid | $31.61 \pm 3.12$ | $0.06 \pm 0.00$ | - | $0.87 \pm 0.08$ |
| 2) External obstacle | $63.91 \pm 5.41$ | $0.08 \pm 0.01$ | - | $1.41 \pm 0.03$ |
| 3) Internal obstacles (removed mega-cells) | $11.41 \pm 2.79$ | $0.05 \pm 0.00$ | - | $1.21 \pm 0.06$ |
| 4) Multiple disconnected grid regions | $35.97 \pm 14.32$ | $0.07 \pm 0.01$ | - | $0.91 \pm 0.02$ |
| 5) Removal of tree mega-cells | $40.37 \pm 10.81$ | $0.07 \pm 0.01$ | $93.09 \pm 20.10$ | $1.12 \pm 0.04$ |

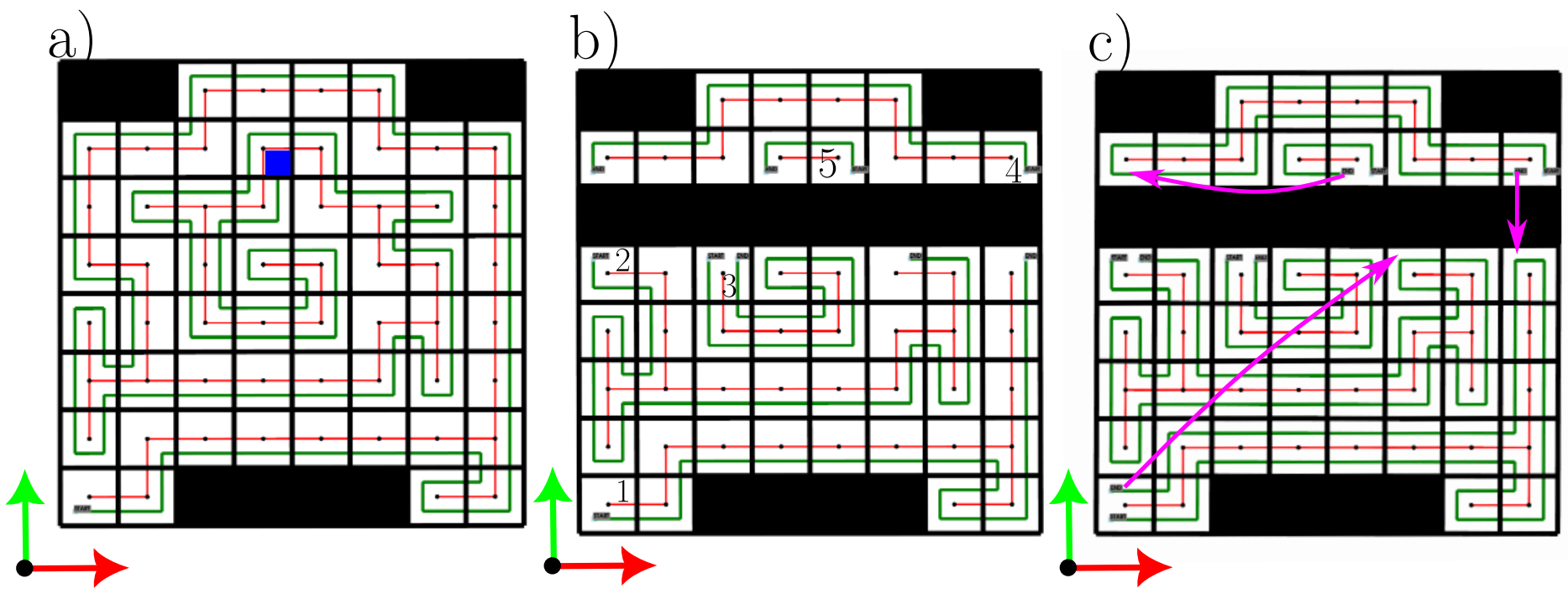

Fig. 5. Coverage paths in the case where the tree and its associated path in a) is split into multiple components due to the removal of the mega-cells in the sixth row shown in black in b). In c), we see the completed path associated with each tree and the transitions between trees in purple.